\documentclass[runningheads]{llncs}
\usepackage{graphicx}

\usepackage{amsmath}
\usepackage{caption}
\usepackage{amsfonts}
\begin{document}

\title{Conditional Infilling GANs for Data Augmentation in Mammogram Classification}

\author{Eric Wu* \inst{1,2}, Kevin Wu* \inst{1,2}, David Cox \inst{1}, William Lotter \inst{1,2}}
\authorrunning{Wu et. al}
\institute{Harvard University, Cambridge, MA \and DeepHealth, Inc., Boston, MA}

\maketitle             
\begin{abstract}
Deep learning approaches to breast cancer detection in mammograms have recently shown promising results. However, such models are constrained by the limited size of publicly available mammography datasets, in large part due to privacy concerns and the high cost of generating expert annotations. Limited dataset size is further exacerbated by substantial class imbalance since ``normal'' images dramatically outnumber those with findings. Given the rapid progress of generative models in synthesizing realistic images, and the known effectiveness of simple data augmentation techniques (e.g. horizontal flipping), we ask if it is possible to synthetically augment mammogram datasets using generative adversarial networks (GANs). We train a class-conditional GAN to perform contextual in-filling, which we then use to synthesize lesions onto healthy screening mammograms. First, we show that GANs are capable of generating high-resolution synthetic mammogram patches. 
Next, we experimentally evaluate using the augmented dataset to improve breast cancer classification performance.
We observe that a ResNet-50 classifier trained with GAN-augmented training data produces a higher AUROC compared to the same model trained only on traditionally augmented data, demonstrating the potential of our approach.

\end{abstract}

\begin{figure}[!h]
\begin{center}
   \includegraphics[width=.75\linewidth]{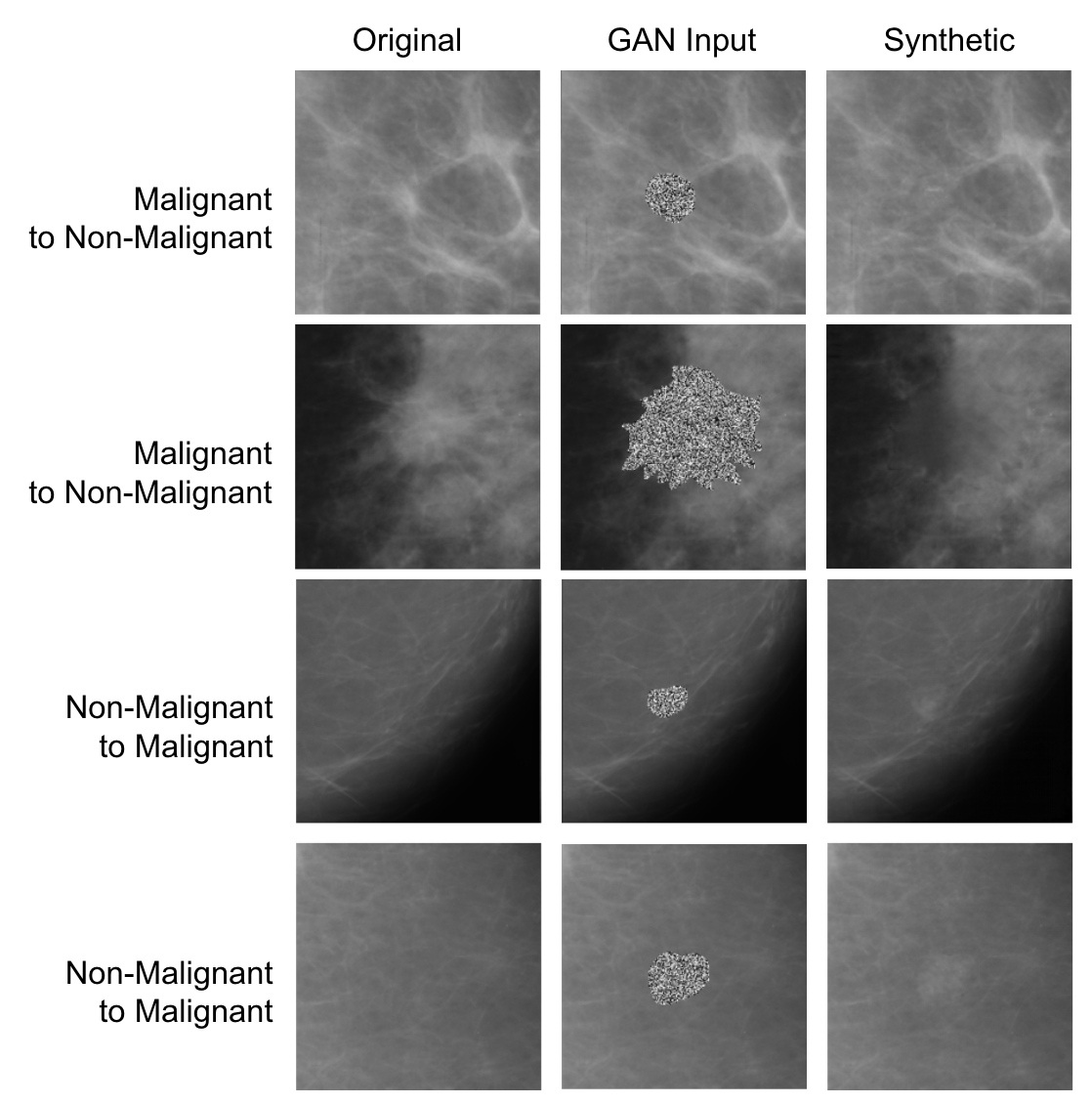}
\end{center}
   \caption{Generated samples from ciGAN using previously unseen patches as context. Each row contains (from left to right) the original image, the input to ciGAN, and the synthetic example generated for the opposite class. 
   The first two rows contain examples of the GAN synthesizing a non-malignant patch from a malignant lesion.
   The third and fourth rows are examples of the GAN synthesizing a malignant lesion on a non-malignant patch, using randomly selected segmentations from other malignant patches. We observe that the GAN is able to incorporate contextual information to smooth out borders of the segmentation masks.
}
\vspace*{-\baselineskip}
\label{fig:mass_gen}
\end{figure}

\section{Introduction}

A major enabler of the recent success of deep learning in computer vision has been the availability of massive-scale, labeled training sets (e.g. ImageNet \cite{russakovsky2015imagenet}).
However, in many medical imaging domains, collecting such datasets is difficult or impossible due to privacy restrictions, the need for expert annotators, and the distribution of data across many sites that cannot share data. 
The class imbalance naturally present in many medical domains, where ``normal'' images dramatically outnumber those with findings, further exacerbates these issues.
{\let\thefootnote\relax\footnote{{*Denotes equal contribution.}}}
A common technique used to combat overfitting is to synthetically increase the size of a dataset through data augmentation, where affine transformations such as flipping or resizing are applied to training images.
The success of these simple techniques raises the question of whether one can further augment training sets using more sophisticated methods.
One potential avenue could be to synthetically generate new training examples altogether.
While generating training samples may seem counterintuitive, rapid progress in designing generative models (particularly generative adversarial networks (GANs) \cite{goodfellowgenerative,gulrajani2017improved,berthelot2017began}) to synthesize highly realistic images merits exploration of this proposal.
Indeed, GANs have been used for data augmentation in several recent works \cite{peng2018jointly,yu2017semantic,wang2017fast,wang2018low,antoniou2017data}, and  investigators have applied GANs to medical images such as magnetic resonance (MR) and computed tomography (CT) \cite{wolterink2017deep,nie2017medical}. Similarly, GANs have been used for data augmentation in liver lesions \cite{frid2018synthetic}, retinal fundi \cite{guibas2017synthetic}, histopathology \cite{hou2017unsupervised}, and chest x-rays \cite{salehinejad2017generalization}.

A particular domain where GANs could be highly effective for data augmentation is cancer detection in mammograms.
The localized nature of many tumors in otherwise seemingly normal tissue suggests a straightforward, first-order procedure for data augmentation: sample a location in a normal mammogram and synthesize a lesion in this location.
This approach also confers benefits to the generative model, as only a smaller patch of the whole image needs to be augmented.
GANs for data augmentation in mammograms is especially promising because of 1) the lack of large-scale public datasets, 2) the small proportion of malignant outcomes in a normal population ($\sim$0.5\%) \cite{breast_2016} and, most importantly, 3) the clinical impact of screening initiatives, with the potential for machine learning to improve quality of care and global population coverage \cite{ribli2018detecting}.

Here, we take a first step towards harnessing GAN-based data augmentation for increasing cancer classification performance in mammography. First, we demonstrate that our GAN architecture (ciGAN) is able to generate a diverse set of synthetic image patches at a high resolution (256x256 pixels). Second, we provide an empirical study on the effectiveness of GAN-based data augmentation for breast cancer classification. Our results indicate that GAN-based augmentation improves mammogram patch-based classification by 0.014 AUC over the baseline model and 0.009 AUC over traditional augmentation techniques alone. 

\begin{figure}[t]
\begin{center}
   \includegraphics[width=0.8\linewidth]{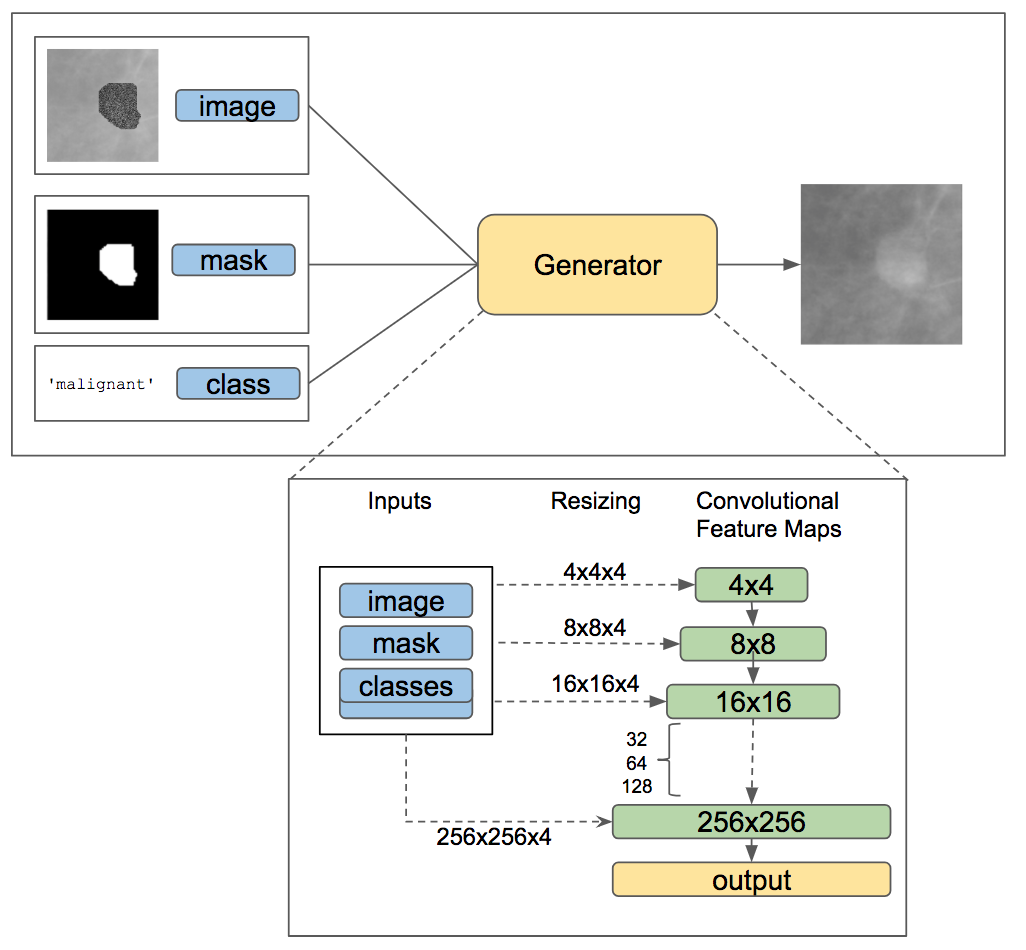}
\end{center}
	\vspace*{-\baselineskip}
   \caption{The ciGAN generator architecture. The inputs consist of four channels (in blue): one context image (where the lesion is replaced with a random noise mask), one lesion mask, and two class channels for indicating a malignant or non-malignant label. Each convolutional block (in green) represents two convolutional layers with an upsampling operation.
}
\vspace*{-1.5\baselineskip}
\label{fig:cigan-gen}
\end{figure}

\section{Proposed Approach: Conditional Infilling GAN}
GANs are known to suffer from convergence issues, especially with high dimensional images \cite{salimans2016improved,gulrajani2017improved,berthelot2017began,kodali2017train}. To address this issue, we construct a GAN using a multi-scale generator architecture trained to infill a segmented area in a target image.
First, our generator is based on a cascading refinement network \cite{chen2017photographic}, where features are generated at multiple scales before being concatenated to improve stability at high resolutions. Second, rather than requiring the generator to replicate redundant context in a mammography patch, we constrain the generator to infill only the segmented lesion (either a mass or calcification). Finally, we use a conditional GAN structure to share learned features between non-malignant and malignant cases \cite{mirza2014conditional}.

\subsection{Architecture}
Our conditional infilling GAN architecture (here on referred to as ciGAN) is outlined in Figure \ref{fig:cigan-gen}. The input is a concatenated stack (in blue) of one grayscale channel with the lesion replaced with uniformly random values between 0 and 1 (the corrupted image), one channel with ones representing the location of the lesion and zeros elsewhere (the mask), and two channels with values as [1,0] representing the non-malignant class or [0,1] as the malignant class (the class labels).
The input stack is downsampled to 4x4 and passed into the first convolutional block (in green), which contains two convolutional layers with 3x3 kernels and ReLU activation functions. The output of this block is upsampled to twice the current resolution (8x8) and then concatenated with an input stack resized to 8x8 before being passed into the second convolutional block. This process is repeated until a final resolution of 256x256 is obtained. The convolutional layers have 128, 128, 64, 64, 32, 32, and 32 kernels from the first to the last block. We use the nearest neighbors method for upsampling. 

The discriminator network has a similar but inverse structure. The input consists of a 256x256 image. This is passed through a convolutional layer with 32 kernels, 3x3 kernel size, and the LeakyReLU \cite{xu2015empirical} activation function, followed by a 2x2 max pooling operation. We apply a total of 5 convolutional layers, doubling the number of kernels each time until the final layer of 512 kernels. This layer is then flattened and passed into a fully connected layer with one unit and a sigmoid activation function.

\subsection{Training Details}
\noindent\textbf{Patch-level training:} Given that most lesions are present within a localized area much smaller than the whole breast image (though context \& global features may also be important), we focus on generating patches (256x256) containing such lesions. This allows us to more meaningfully measure the effects of GAN-augmented training as opposed to using the whole image. Furthermore, patch-level pre-training has been shown to increase generalization for full images \cite{lotter2017multi,therapixel,shen2017end}.

The ciGAN model is trained using a combination of the following three loss functions:

\noindent\textbf{Feature Loss:} For a feature loss, we utilize the VGG-19 \cite{simonyan2014very} convolutional neural network, pre-trained on the ImageNet dataset. Real and generated images are passed through the network to extract the feature maps at the $\text{pool}_1$, $\text{pool}_2$, and $\text{pool}_3$ layers, where the mean of the absolute errors is taken between the maps. This loss encourages the features of the generator to match the real image at different spatial resolutions and feature complexities. Letting ${\Phi_i}$ be the collection of layers in $\Phi$, the VGG19 network, where $\Phi_0$ is the input image, we define VGG loss for the real image R and generated image S as:

$$
\mathcal{L}_{R,S}(\theta) = \sum_l || \Phi_l(R) - \Phi_l(S)||_1
$$

\noindent\textbf{Adversarial Loss:} We use the adversarial loss formulated in \cite{goodfellow2014generative}, which seeks optimize over the following mini-max game involving generator G and discriminator D:

$$
\min_{G} \max_{D} \mathcal{L}_{GAN}(G,D)
$$
$$
\mathcal{L}_{GAN}(G,D) = \mathbb{E}_{(c,R)}[\log D(c,R)]+\mathbb{E}_{R}[\log(1-D(c, S)]
$$
Where $c$ is the class label, $R$ is a real image, and $S$ is the generated image.

\noindent\textbf{Boundary Loss:} To encourage smoothing between the infilled component and the context of a generated image, we introduce a boundary loss, which is the $L_{1}$ difference between the real and generated image at the boundary:
$$
B_{R,S}(\theta) = || w \odot (R - S)||_1
$$
Where R is the real image, S is the generated image, $w$ is the mask boundary with a Gaussian filter of standard deviation 10 applied, and $\odot$ is the element-wise product.

\noindent\textbf{Training details:} In our implementation, we alternate between training the generator and discriminator when the loss for either drops below 0.3. We use the Adam \cite{kingma2014adam} optimizer with $\beta_1$=0.9, $\beta_2$=0.999, $\epsilon = 10^{-8}$, a learning rate of 1e-4, and batch size of 8. To stabilize training, we first pre-train the generator exclusively on feature loss for 10,000 iterations. Then, we train the generator and discriminator on all losses for an additional 100,000 iterations. We weigh each loss with coefficients 1.0, 10.0, and 10000.0 for GAN loss, feature loss, and boundary loss, respectively.
 
\section{Experiments}

\subsection{DDSM Dataset}

The DDSM (Digital Database for Screening Mammography) dataset contains 10,480 total images, with 1,832 (17.5\%) malignant cases and 8,648 (82.5\%) non-malignant cases. Image patches are labeled as malignant or non-malignant along with the segmentation masks in the dataset. Both calcifications and masses are used and non-malignant patches contain both benign and non-lesion patches.

We apply a 80\% training, 10\% validation, and 10\% testing split on the dataset. To process full resolution images into patches, we take each image ($\sim$5500x3000 pixels) and resize to a target range of 1375x750 while ensuring the original aspect ratio is maintained, as described in \cite{lotter2017multi}. For both non-malignant and malignant cases, we generate 100,000 random 256x256 pixel patches and only accept patches that consist of more than 75\% breast tissue.

\subsection{GAN-based data augmentation}
We evaluate the effectiveness of GAN-based data augmentation on the task of cancer detection. We choose the ResNet-50 architecture as our classifier network \cite{he2016deep}. We use the Adam optimizer with an initial learning rate of $10^{-5}$ and $\beta_1$=0.9, $\beta_2$=0.999, $\epsilon = 10^{-8}$. To achieve better performance, we initialize the classifier with ImageNet weights. For each regime, we train for 10,000 iterations on a batch size of 32 with a 0.9 learning rate decay rate every 2,000 iterations. The GAN is only trained on the training data used for the classifier.

For traditional image data augmentation, we use random rotations up to 30 degrees, horizontal flipping, and rescaling by a factor between 0.75 and 1.25.
For augmentation with ciGAN, we double our existing dataset via the following procedure: for each non-malignant image, we generate a malignant lesion onto it using a mask from another malignant lesion. For each malignant patch, we remove the malignant lesion and generate a non-malignant image in its place. In total, we produce 8,648 synthetically generated malignant patches and 1,832 synthetically generated non-malignant patches.
We train the classifier by initially training on equal proportions of real and synthetic data. Every 1000 iterations, we increase the relative proportion of real data used by 20\%, such that the final iteration is trained on 90\% real data. We observe that this regime helps prevent early overfitting and greater generalization for later epochs.

\begin{table}[t]
\centering
\renewcommand{\arraystretch}{1.5}
\begin{tabular}{|c|c|}
\hline
\textbf{ Data augmentation scheme } & \textbf{ AUC } \\ \hline
Baseline (no augmentation)        & 0.882                         \\ \hline
Traditional augmentation          & 0.887                         \\ \hline
ciGAN + Traditional aug         & \textbf{0.896}                             \\ \hline
\end{tabular}
\caption{ROC AUC (Area 
under ROC curve) for three augmentation schemes.}
\label{table:aug-results}
\end{table}

\subsection{Results}
Table \ref{table:aug-results} contains the results of three classification experiments. ciGAN, combined with traditional augmentation, achieves an AUC of $0.896$. This outperforms  the baseline (no augmentation) model by 0.014 AUC ($p<0.01$, DeLong method \cite{delong1988comparing}) and traditional augmentation model by 0.009 AUC ($p<0.05$). Direct comparison of our results with similar works is difficult given that DDSM does not have standardized training/testing splits, but we find that our models compare on par or favorably to other DDSM patch classification efforts \cite{shen2017end,zhu2017deep,levy2016breast}.
 
\section{Conclusion}

Recent efforts for using deep learning for cancer detection in mammograms have yielded promising results. 
One major limiting factor for continued progress is the scarcity of data, and especially cancer positive exams. 
Given the success of simple data augmentation techniques and the recent progress in generative adversarial networks (GANs), we ask whether GANs can be used to synthetically increase the size of training data by generating examples of mammogram lesions. 
We employ a multi-scale class-conditional GAN with mask infilling (ciGAN), and demonstrate that our GAN indeed is able to generate realistic lesions, which improves subsequent classification performance above traditional augmentation techniques.
ciGAN addresses critical issues in other GAN architectures, such as training instability and resolution detail. 
Scarcity of data and class imbalance are common constraints in medical imaging tasks, and we believe our techniques can help address these issues in a variety of settings.

\noindent\textbf{Acknowledgements:} This work was supported by the National Science Foundation (NSF IIS 1409097).
\vspace{-\baselineskip}
  \bibliographystyle{ieeetr}
  \bibliography{references}

\begin{thebibliography}{10}

\bibitem{russakovsky2015imagenet}
O.~Russakovsky, J.~Deng, H.~Su, J.~Krause, S.~Satheesh, S.~Ma, Z.~Huang,
  A.~Karpathy, A.~Khosla, M.~Bernstein, {\em et~al.}, ``Imagenet large scale
  visual recognition challenge,'' {\em IJCV}, 2015.

\bibitem{goodfellowgenerative}
I.~J. Goodfellow, J.~Pouget-Abadie, M.~Mirza, B.~Xu, D.~Warde-Farley, S.~Ozair,
  A.~Courville, and Y.~Bengio, ``Generative adversarial nets,''

\bibitem{gulrajani2017improved}
I.~Gulrajani, F.~Ahmed, M.~Arjovsky, V.~Dumoulin, and A.~C. Courville,
  ``Improved training of wasserstein gans,'' in {\em NIPS}, pp.~5769--5779,
  2017.

\bibitem{berthelot2017began}
D.~Berthelot, T.~Schumm, and L.~Metz, ``Began: Boundary equilibrium generative
  adversarial networks,'' {\em arXiv preprint arXiv:1703.10717}, 2017.

\bibitem{peng2018jointly}
X.~Peng, Z.~Tang, F.~Yang, R.~S. Feris, and D.~Metaxas, ``Jointly optimize data
  augmentation and network training: Adversarial data augmentation in human
  pose estimation,'' in {\em CVPR}, 2018.

\bibitem{yu2017semantic}
A.~Yu and K.~Grauman, ``Semantic jitter: Dense supervision for visual
  comparisons via synthetic images,'' tech. rep., Technical Report, 2017.

\bibitem{wang2017fast}
X.~Wang, A.~Shrivastava, and A.~Gupta, ``A-fast-rcnn: Hard positive generation
  via adversary for object detection,'' {\em arXiv}, vol.~2, 2017.

\bibitem{wang2018low}
Y.-X. Wang, R.~Girshick, M.~Hebert, and B.~Hariharan, ``Low-shot learning from
  imaginary data,'' {\em arXiv preprint arXiv:1801.05401}, 2018.

\bibitem{antoniou2017data}
A.~Antoniou, A.~Storkey, and H.~Edwards, ``Data augmentation generative
  adversarial networks,'' {\em arXiv preprint arXiv:1711.04340}, 2017.

\bibitem{wolterink2017deep}
J.~M. Wolterink, A.~M. Dinkla, M.~H. Savenije, P.~R. Seevinck, C.~A. van~den
  Berg, and I.~I{\v{s}}gum, ``Deep mr to ct synthesis using unpaired data,'' in
  {\em International Workshop on Simulation and Synthesis in Medical Imaging},
  Springer, 2017.

\bibitem{nie2017medical}
D.~Nie, R.~Trullo, J.~Lian, C.~Petitjean, S.~Ruan, Q.~Wang, and D.~Shen,
  ``Medical image synthesis with context-aware generative adversarial
  networks,'' in {\em MICCAI}, pp.~417--425, Springer, 2017.

\bibitem{frid2018synthetic}
M.~Frid-Adar, E.~Klang, M.~Amitai, J.~Goldberger, and H.~Greenspan, ``Synthetic
  data augmentation using gan for improved liver lesion classification,'' {\em
  arXiv preprint arXiv:1801.02385}, 2018.

\bibitem{guibas2017synthetic}
J.~T. Guibas, T.~S. Virdi, and P.~S. Li, ``Synthetic medical images from dual
  generative adversarial networks,'' {\em arXiv preprint arXiv:1709.01872},
  2017.

\bibitem{hou2017unsupervised}
L.~Hou, A.~Agarwal, D.~Samaras, T.~M. Kurc, R.~R. Gupta, and J.~H. Saltz,
  ``Unsupervised histopathology image synthesis,'' {\em arXiv}, 2017.

\bibitem{salehinejad2017generalization}
H.~Salehinejad, S.~Valaee, T.~Dowdell, E.~Colak, and J.~Barfett,
  ``Generalization of deep neural networks for chest pathology classification
  in x-rays using generative adversarial networks,'' {\em arXiv preprint
  arXiv:1712.01636}, 2017.

\bibitem{breast_2016}
Cancer.gov, ``Cancer facts and figures, 2015-2016,'' 2016.

\bibitem{ribli2018detecting}
D.~Ribli, A.~Horv{\'a}th, Z.~Unger, P.~Pollner, and I.~Csabai, ``Detecting and
  classifying lesions in mammograms with deep learning,'' {\em Scientific
  reports}, vol.~8, no.~1, p.~4165, 2018.

\bibitem{salimans2016improved}
T.~Salimans, I.~Goodfellow, W.~Zaremba, V.~Cheung, A.~Radford, and X.~Chen,
  ``Improved techniques for training gans,'' in {\em NIPS}, pp.~2234--2242,
  2016.

\bibitem{kodali2017train}
N.~Kodali, J.~Abernethy, J.~Hays, and Z.~Kira, ``How to train your dragan,''
  {\em arXiv preprint arXiv:1705.07215}, 2017.

\bibitem{chen2017photographic}
Q.~Chen and V.~Koltun, ``Photographic image synthesis with cascaded refinement
  networks,'' in {\em ICCV 2017}, pp.~1520--1529, IEEE, 2017.

\bibitem{mirza2014conditional}
M.~Mirza and S.~Osindero, ``Conditional generative adversarial nets,'' {\em
  arXiv preprint arXiv:1411.1784}, 2014.

\bibitem{xu2015empirical}
B.~Xu, N.~Wang, T.~Chen, and M.~Li, ``Empirical evaluation of rectified
  activations in convolutional network,'' {\em arXiv}, 2015.

\bibitem{lotter2017multi}
W.~Lotter, G.~Sorensen, and D.~Cox, ``A multi-scale cnn and curriculum learning
  strategy for mammogram classification,'' in {\em Deep Learning in Medical
  Image Analysis and Multimodal Learning for Clinical Decision Support},
  Springer, 2017.

\bibitem{therapixel}
Y.~Nikulin, ``Dm challenge yaroslav nikulin (therapixel),'' {\em Synapse.org},
  2017.

\bibitem{shen2017end}
L.~Shen, ``End-to-end training for whole image breast cancer diagnosis using an
  all convolutional design,'' {\em arXiv preprint arXiv:1708.09427}, 2017.

\bibitem{simonyan2014very}
K.~Simonyan and A.~Zisserman, ``Very deep convolutional networks for
  large-scale image recognition,'' {\em arXiv preprint arXiv:1409.1556}, 2014.

\bibitem{goodfellow2014generative}
I.~Goodfellow, J.~Pouget-Abadie, M.~Mirza, B.~Xu, D.~Warde-Farley, S.~Ozair,
  A.~Courville, and Y.~Bengio, ``Generative adversarial nets,'' in {\em NIPS}.

\bibitem{kingma2014adam}
D.~P. Kingma and J.~Ba, ``Adam: A method for stochastic optimization,'' {\em
  arXiv preprint arXiv:1412.6980}, 2014.

\bibitem{he2016deep}
K.~He, X.~Zhang, S.~Ren, and J.~Sun, ``Deep residual learning for image
  recognition,'' in {\em CVPR}, pp.~770--778, 2016.

\bibitem{delong1988comparing}
E.~R. DeLong, D.~M. DeLong, and D.~L. Clarke-Pearson, ``Comparing the areas
  under two or more correlated receiver operating characteristic curves: a
  nonparametric approach,'' {\em Biometrics}, pp.~837--845, 1988.

\bibitem{zhu2017deep}
W.~Zhu, Q.~Lou, Y.~S. Vang, and X.~Xie, ``Deep multi-instance networks with
  sparse label assignment for whole mammogram classification,'' in {\em
  MICCAI}, 2017.

\bibitem{levy2016breast}
D.~L{\'e}vy and A.~Jain, ``Breast mass classification from mammograms using
  deep convolutional neural networks,'' {\em arXiv}, 2016.

\end{thebibliography}


@misc{CancerGov, url={https://seer.cancer.gov/}, journal={Surveillance, Epidemiology, and End Results Program}}
@article{delong1988comparing,
  title={Comparing the areas under two or more correlated receiver operating characteristic curves: a nonparametric approach},
  author={DeLong, Elizabeth R and DeLong, David M and Clarke-Pearson, Daniel L},
  journal={Biometrics},
  pages={837--845},
  year={1988},
  publisher={JSTOR}
}
@inproceedings{zhu2017deep,
  title={Deep multi-instance networks with sparse label assignment for whole mammogram classification},
  author={Zhu, Wentao and Lou, Qi and Vang, Yeeleng Scott and Xie, Xiaohui},
  booktitle={MICCAI},
  year={2017}
}
@article{levy2016breast,
  title={Breast mass classification from mammograms using deep convolutional neural networks},
  author={L{\'e}vy, Daniel and Jain, Arzav},
  journal={arXiv},
  year={2016}
}
@inproceedings{ball2007digital,
  title={Digital mammographic computer aided diagnosis (cad) using adaptive level set segmentation},
  author={Ball, John E and Bruce, Lori Mann},
  booktitle={Engineering in Medicine and Biology Society, 2007. EMBS 2007. 29th Annual International Conference of the IEEE},
  year={2007},
  organization={IEEE}
}
@article{xu2015empirical,
  title={Empirical evaluation of rectified activations in convolutional network},
  author={Xu, Bing and Wang, Naiyan and Chen, Tianqi and Li, Mu},
  journal={arXiv},
  year={2015}
}

@article{varela2006use,
  title={Use of border information in the classification of mammographic masses},
  author={Varela, C and Timp, S and Karssemeijer, N},
  journal={Physics in medicine \& biology},
  year={2006}
}

@article{therapixel, author={Nikulin, Yaroslav},  title={DM Challenge Yaroslav Nikulin (Therapixel)}, journal={Synapse.org}, year={2017}}

@article{shen2017end,
  title={End-to-end Training for Whole Image Breast Cancer Diagnosis using An All Convolutional Design},
  author={Shen, Li},
  journal={arXiv preprint arXiv:1708.09427},
  year={2017}
}

@misc{breast_2016, author={Cancer.gov}, title={Cancer Facts and Figures, 2015-2016}, journal={American Cancer Society}, year={2016}}

@article{elmore2009variability,
  title={Variability in interpretive performance at screening mammography and radiologists’ characteristics associated with accuracy},
  author={Elmore, Joann G and Jackson, Sara L and Abraham, Linn and Miglioretti, Diana L and Carney, Patricia A and Geller, Berta M and Yankaskas, Bonnie C and Kerlikowske, Karla and Onega, Tracy and Rosenberg, Robert D and others},
  journal={Radiology},
  year={2009}
}
@incollection{lotter2017multi,
  title={A multi-scale cnn and curriculum learning strategy for mammogram classification},
  author={Lotter, William and Sorensen, Greg and Cox, David},
  booktitle={Deep Learning in Medical Image Analysis and Multimodal Learning for Clinical Decision Support},
  year={2017},
  publisher={Springer}
}

@article{lehman2016national,
  title={National performance benchmarks for modern screening digital mammography: update from the Breast Cancer Surveillance Consortium},
  author={Lehman, Constance D and Arao, Robert F and Sprague, Brian L and Lee, Janie M and Buist, Diana SM and Kerlikowske, Karla and Henderson, Louise M and Onega, Tracy and Tosteson, Anna NA and Rauscher, Garth H and others},
  journal={Radiology},
  year={2016},
  publisher={Radiological Society of North America}
}

@article{nishikawa2007current,
  title={Current status and future directions of computer-aided diagnosis in mammography},
  author={Nishikawa, Robert M},
  journal={Computerized Medical Imaging and Graphics},
  year={2007},
  publisher={Elsevier}
}

@article{russakovsky2015imagenet,
  title={Imagenet large scale visual recognition challenge},
  author={Russakovsky, Olga and Deng, Jia and Su, Hao and Krause, Jonathan and Satheesh, Sanjeev and Ma, Sean and Huang, Zhiheng and Karpathy, Andrej and Khosla, Aditya and Bernstein, Michael and others},
  journal={IJCV},
  year={2015},
  publisher={Springer}
}

@article{goodfellowgenerative,
  title={Generative Adversarial Nets},
  author={Goodfellow, Ian J and Pouget-Abadie, Jean and Mirza, Mehdi and Xu, Bing and Warde-Farley, David and Ozair, Sherjil and Courville, Aaron and Bengio, Yoshua}
}

@article{heath2000digital,
  title={The digital database for screening mammography},
  author={Heath, M and Bowyer, K and Kopans, D and Moore, R}
}

@inproceedings{chen2017photographic,
  title={Photographic Image Synthesis with Cascaded Refinement Networks},
  author={Chen, Qifeng and Koltun, Vladlen},
  booktitle={ICCV 2017},
  pages={1520--1529},
  year={2017},
  organization={IEEE}
}

@inproceedings{wolterink2017deep,
  title={Deep MR to CT synthesis using unpaired data},
  author={Wolterink, Jelmer M and Dinkla, Anna M and Savenije, Mark HF and Seevinck, Peter R and van den Berg, Cornelis AT and I{\v{s}}gum, Ivana},
  booktitle={International Workshop on Simulation and Synthesis in Medical Imaging},
  year={2017},
  organization={Springer}
}

@inproceedings{peng2018jointly,
  title={Jointly optimize data augmentation and network training: Adversarial data augmentation in human pose estimation},
  author={Peng, Xi and Tang, Zhiqiang and Yang, Fei and Feris, Rogerio S and Metaxas, Dimitris},
  booktitle={CVPR},
  year={2018}
}

@techreport{yu2017semantic,
  title={Semantic jitter: Dense supervision for visual comparisons via synthetic images},
  author={Yu, Aron and Grauman, Kristen},
  year={2017},
  institution={Technical Report}
}

@article{wang2017fast,
  title={A-fast-rcnn: Hard positive generation via adversary for object detection},
  author={Wang, Xiaolong and Shrivastava, Abhinav and Gupta, Abhinav},
  journal={arXiv},
  volume={2},
  year={2017}
}

@article{antoniou2017data,
  title={Data Augmentation Generative Adversarial Networks},
  author={Antoniou, Antreas and Storkey, Amos and Edwards, Harrison},
  journal={arXiv preprint arXiv:1711.04340},
  year={2017}
}

@article{wang2018low,
  title={Low-Shot Learning from Imaginary Data},
  author={Wang, Yu-Xiong and Girshick, Ross and Hebert, Martial and Hariharan, Bharath},
  journal={arXiv preprint arXiv:1801.05401},
  year={2018}
}


@article{arevalo2016representation,
  title={Representation learning for mammography mass lesion classification with convolutional neural networks},
  author={Arevalo, John and Gonz{\'a}lez, Fabio A and Ramos-Poll{\'a}n, Ra{\'u}l and Oliveira, Jose L and Lopez, Miguel Angel Guevara},
  journal={Computer methods and programs in biomedicine},
  volume={127},
  pages={248--257},
  year={2016},
  publisher={Elsevier}
}

@inproceedings{mordang2016automatic,
  title={Automatic microcalcification detection in multi-vendor mammography using convolutional neural networks},
  author={Mordang, Jan-Jurre and Janssen, Tim and Bria, Alessandro and Kooi, Thijs and Gubern-M{\'e}rida, Albert and Karssemeijer, Nico},
  booktitle={International Workshop on Digital Mammography},
  pages={35--42},
  year={2016},
  organization={Springer}
}

@inproceedings{carneiro2015unregistered,
  title={Unregistered multiview mammogram analysis with pre-trained deep learning models},
  author={Carneiro, Gustavo and Nascimento, Jacinto and Bradley, Andrew P},
  booktitle={MICCAI},
  pages={652--660},
  year={2015},
  organization={Springer}
}

@inproceedings{dhungel2015automated,
  title={Automated mass detection in mammograms using cascaded deep learning and random forests},
  author={Dhungel, Neeraj and Carneiro, Gustavo and Bradley, Andrew P},
  booktitle={Digital Image Computing: Techniques and Applications (DICTA), 2015 International Conference on},
  pages={1--8},
  year={2015},
  organization={IEEE}
}

@inproceedings{dhungel2015deep,
  title={Deep structured learning for mass segmentation from mammograms},
  author={Dhungel, Neeraj and Carneiro, Gustavo and Bradley, Andrew P},
  booktitle={Image Processing (ICIP), 2015 IEEE International Conference on},
  pages={2950--2954},
  year={2015},
  organization={IEEE}
}

@article{geras2017high,
  title={High-resolution breast cancer screening with multi-view deep convolutional neural networks},
  author={Geras, Krzysztof J and Wolfson, Stacey and Shen, Yiqiu and Kim, S and Moy, Linda and Cho, Kyunghyun},
  journal={arXiv preprint arXiv:1703.07047},
  year={2017}
}

@article{kooi2017large,
  title={Large scale deep learning for computer aided detection of mammographic lesions},
  author={Kooi, Thijs and Litjens, Geert and van Ginneken, Bram and Gubern-M{\'e}rida, Albert and S{\'a}nchez, Clara I and Mann, Ritse and den Heeten, Ard and Karssemeijer, Nico},
  journal={Medical image analysis},
  volume={35},
  pages={303--312},
  year={2017},
  publisher={Elsevier}
}

@article{levy2016breast,
  title={Breast mass classification from mammograms using deep convolutional neural networks},
  author={L{\'e}vy, Daniel and Jain, Arzav},
  journal={arXiv preprint arXiv:1612.00542},
  year={2016}
}

@article{yi2017optimizing,
  title={Optimizing and visualizing deep learning for benign/malignant classification in breast tumors},
  author={Yi, Darvin and Sawyer, Rebecca Lynn and Cohn III, David and Dunnmon, Jared and Lam, Carson and Xiao, Xuerong and Rubin, Daniel},
  journal={arXiv preprint arXiv:1705.06362},
  year={2017}
}

@inproceedings{zhu2017deep,
  title={Deep multi-instance networks with sparse label assignment for whole mammogram classification},
  author={Zhu, Wentao and Lou, Qi and Vang, Yeeleng Scott and Xie, Xiaohui},
  booktitle={MICCAI},
  pages={603--611},
  year={2017}
}

@inproceedings{he2016deep,
  title={Deep residual learning for image recognition},
  author={He, Kaiming and Zhang, Xiangyu and Ren, Shaoqing and Sun, Jian},
  booktitle={CVPR},
  pages={770--778},
  year={2016}
}


@article{,
  title={Conditional generative adversarial nets},
  author={Mirza, Mehdi and Osindero, Simon},
  journal={arXiv preprint arXiv:1411.1784},
  year={2014}
}

@article{hou2017unsupervised,
  title={Unsupervised Histopathology Image Synthesis},
  author={Hou, Le and Agarwal, Ayush and Samaras, Dimitris and Kurc, Tahsin M and Gupta, Rajarsi R and Saltz, Joel H},
  journal={arXiv},
  year={2017}
}
@article{frid2018synthetic,
  title={Synthetic Data Augmentation using GAN for Improved Liver Lesion Classification},
  author={Frid-Adar, Maayan and Klang, Eyal and Amitai, Michal and Goldberger, Jacob and Greenspan, Hayit},
  journal={arXiv preprint arXiv:1801.02385},
  year={2018}
}

@article{costa2017towards,
  title={Towards adversarial retinal image synthesis},
  author={Costa, Pedro and Galdran, Adrian and Meyer, Maria In{\^e}s and Abr{\`a}moff, Michael David and Niemeijer, Meindert and Mendon{\c{c}}a, Ana Maria and Campilho, Aur{\'e}lio},
  journal={arXiv preprint arXiv:1701.08974},
  year={2017}
}

@inproceedings{nie2017medical,
  title={Medical image synthesis with context-aware generative adversarial networks},
  author={Nie, Dong and Trullo, Roger and Lian, Jun and Petitjean, Caroline and Ruan, Su and Wang, Qian and Shen, Dinggang},
  booktitle={MICCAI},
  pages={417--425},
  year={2017},
  organization={Springer}
}

@misc{CDC_stat, title={National Center for Health Statistics}, url={https://www.cdc.gov/nchs/fastats/mammography.htm}, journal={Centers for Disease Control and Prevention}, publisher={Centers for Disease Control and Prevention}, year={2017}, month={May}}

@inproceedings{oquab2014learning,
  title={Learning and transferring mid-level image representations using convolutional neural networks},
  author={Oquab, Maxime and Bottou, Leon and Laptev, Ivan and Sivic, Josef},
  booktitle={CVPR, 2014 IEEE Conference on},
  pages={1717--1724},
  year={2014},
  organization={IEEE}
}
@article{karras2017progressive,
  title={Progressive growing of gans for improved quality, stability, and variation},
  author={Karras, Tero and Aila, Timo and Laine, Samuli and Lehtinen, Jaakko},
  journal={arXiv preprint arXiv:1710.10196},
  year={2017}
}

@article{litjens2017survey,
  title={A survey on deep learning in medical image analysis},
  author={Litjens, Geert and Kooi, Thijs and Bejnordi, Babak Ehteshami and Setio, Arnaud Arindra Adiyoso and Ciompi, Francesco and Ghafoorian, Mohsen and van der Laak, Jeroen AWM and van Ginneken, Bram and S{\'a}nchez, Clara I},
  journal={Medical image analysis},
  volume={42},
  pages={60--88},
  year={2017},
  publisher={Elsevier}
}

@article{singh2018conditional,
  title={Conditional Generative Adversarial and Convolutional Networks for X-ray Breast Mass Segmentation and Shape Classification},
  author={Singh, Vivek Kumar and Romani, Santiago and Rashwan, Hatem A and Akram, Farhan and Pandey, Nidhi and Sarker, Md and Kamal, Mostafa and Barrena, Jordina Torrents and Saleh, Adel and Arenas, Meritxell and others},
  journal={arXiv preprint arXiv:1805.10207},
  year={2018}
}

@inproceedings{kim2018feasibility,
  title={Feasibility study of deep convolutional generative adversarial networks to generate mammography images},
  author={Kim, Gihun and Shim, Hyunjung and Baek, Jongduk},
  booktitle={Medical Imaging 2018: Image Perception, Observer Performance, and Technology Assessment},
  volume={10577},
  pages={105771C},
  year={2018},
  organization={International Society for Optics and Photonics}
}


@article{nguyen2017long,
  title={Long range iris recognition: A survey},
  author={Nguyen, Kien and Fookes, Clinton and Jillela, Raghavender and Sridharan, Sridha and Ross, Arun},
  journal={Pattern Recognition},
  volume={72},
  pages={123--143},
  year={2017},
  publisher={Elsevier}
}
@inproceedings{he2017mask,
  title={Mask r-cnn},
  author={He, Kaiming and Gkioxari, Georgia and Doll{\'a}r, Piotr and Girshick, Ross},
  booktitle={Computer Vision (ICCV), 2017 IEEE International Conference on},
  pages={2980--2988},
  year={2017},
  organization={IEEE}
}
@article{kingma2013auto,
  title={Auto-encoding variational bayes},
  author={Kingma, Diederik P and Welling, Max},
  journal={arXiv preprint arXiv:1312.6114},
  year={2013}
}
@inproceedings{krizhevsky2012imagenet,
  title={Imagenet classification with deep convolutional neural networks},
  author={Krizhevsky, Alex and Sutskever, Ilya and Hinton, Geoffrey E},
  booktitle={NIPS},
  pages={1097--1105},
  year={2012}
}
@article{lecun1998gradient,
  title={Gradient-based learning applied to document recognition},
  author={LeCun, Yann and Bottou, L{\'e}on and Bengio, Yoshua and Haffner, Patrick},
  journal={Proceedings of the IEEE},
  volume={86},
  number={11},
  pages={2278--2324},
  year={1998},
  publisher={IEEE}
}
@inproceedings{goodfellow2014generative,
  title={Generative adversarial nets},
  author={Goodfellow, Ian and Pouget-Abadie, Jean and Mirza, Mehdi and Xu, Bing and Warde-Farley, David and Ozair, Sherjil and Courville, Aaron and Bengio, Yoshua},
  booktitle={NIPS},},
  year={2014}
}
@misc{besbes_2017, title={Understanding deep Convolutional Neural Networks with a practical use-case in Tensorflow and Keras}, url={https://ahmedbesbes.com/understanding-deep-convolutional-neural-networks-with-a-practical-use-case-in-tensorflow-and-keras.html}, journal={Ahmed BESBES - Data Science Portfolio}, author={Besbes, Ahmed}, year={2017}, month={Nov}}
@misc{catalunya_2016, title={Deep Learning for Computer Vision: Generative models and adversarial ...}, url={https://www.slideshare.net/xavigiro/deep-learning-for-computer-vision-generative-models-and-adversarial-training-upc-2016}, journal={LinkedIn SlideShare}, author={Catalunya, Universitat Politècnica de}, year={2016}, month={Aug}}
@misc{shafkat_2018, title={Intuitively Understanding Variational Autoencoders – Towards Data Science}, url={https://towardsdatascience.com/intuitively-understanding-variational-autoencoders-1bfe67eb5daf}, journal={Towards Data Science}, publisher={Towards Data Science}, author={Shafkat, Irhum}, year={2018}, month={Feb}}
@misc{deepmind, title={Applying machine learning to mammography screening for breast cancer}, author={Deepmind}, url={https://deepmind.com/blog/applying-machine-learning-mammography/}, journal={DeepMind}}
@misc{nci, title={Mammograms Fact Sheet}, author={National Cancer Institute}, url={https://www.cancer.gov/types/breast/mammograms-fact-sheet}, journal={National Cancer Institute}}
@article{jorgensen2009overdiagnosis,
  title={Overdiagnosis in publicly organised mammography screening programmes: systematic review of incidence trends},
  author={J{\o}rgensen, Karsten Juhl and G{\o}tzsche, Peter C},
  journal={Bmj},
  volume={339},
  pages={b2587},
  year={2009},
  publisher={British Medical Journal Publishing Group}
}
@incollection{lotter2017multi,
  title={A multi-scale cnn and curriculum learning strategy for mammogram classification},
  author={Lotter, William and Sorensen, Greg and Cox, David},
  booktitle={Deep Learning in Medical Image Analysis and Multimodal Learning for Clinical Decision Support},
  pages={169--177},
  year={2017},
  publisher={Springer}
}
@inproceedings{he2016identity,
  title={Identity mappings in deep residual networks},
  author={He, Kaiming and Zhang, Xiangyu and Ren, Shaoqing and Sun, Jian},
  booktitle={European Conference on Computer Vision},
  pages={630--645},
  year={2016},
  organization={Springer}
}
@article{ribli2018detecting,
  title={Detecting and classifying lesions in mammograms with Deep Learning},
  author={Ribli, Dezs{\H{o}} and Horv{\'a}th, Anna and Unger, Zsuzsa and Pollner, P{\'e}ter and Csabai, Istv{\'a}n},
  journal={Scientific reports},
  volume={8},
  number={1},
  pages={4165},
  year={2018},
  publisher={Nature Publishing Group}
}
@inproceedings{petersen2014breast,
  title={Breast tissue segmentation and mammographic risk scoring using deep learning},
  author={Petersen, Kersten and Nielsen, Mads and Diao, Pengfei and Karssemeijer, Nico and Lillholm, Martin},
  booktitle={International Workshop on Digital Mammography},
  pages={88--94},
  year={2014},
  organization={Springer}
}
@inproceedings{mordang2016automatic,
  title={Automatic microcalcification detection in multi-vendor mammography using convolutional neural networks},
  author={Mordang, Jan-Jurre and Janssen, Tim and Bria, Alessandro and Kooi, Thijs and Gubern-M{\'e}rida, Albert and Karssemeijer, Nico},
  booktitle={International Workshop on Digital Mammography},
  pages={35--42},
  year={2016},
  organization={Springer}
}
@inproceedings{carneiro2015unregistered,
  title={Unregistered multiview mammogram analysis with pre-trained deep learning models},
  author={Carneiro, Gustavo and Nascimento, Jacinto and Bradley, Andrew P},
  booktitle={MICCAI}},
  pages={652--660},
  year={2015},
  organization={Springer}
}
@article{heath2000digital,
  title={The digital database for screening mammography},
  author={Heath, M and Bowyer, K and Kopans, D and Moore, R and Kegelmeyer, P},
  journal={Digital mammography},
  pages={431--434},
  year={2000}
}
@article{goodfellow2016nips,
  title={NIPS 2016 tutorial: Generative adversarial networks},
  author={Goodfellow, Ian},
  journal={arXiv preprint arXiv:1701.00160},
  year={2016}
}
@misc{uscher, title={Suspicious Mammogram Result: Now What?}, url={https://www.webmd.com/breast-cancer/features/abnormal-mammogram-results#1}, journal={WebMD}, publisher={WebMD}, author={Uscher, Jen}}
@article{mirza2014conditional,
  title={Conditional generative adversarial nets},
  author={Mirza, Mehdi and Osindero, Simon},
  journal={arXiv preprint arXiv:1411.1784},
  year={2014}
}
@inproceedings{chen2017photographic,
  title={Photographic image synthesis with cascaded refinement networks},
  author={Chen, Qifeng and Koltun, Vladlen}
}
@article{wang2017high,
  title={High-Resolution Image Synthesis and Semantic Manipulation with Conditional GANs},
  author={Wang, Ting-Chun and Liu, Ming-Yu and Zhu, Jun-Yan and Tao, Andrew and Kautz, Jan and Catanzaro, Bryan},
  journal={arXiv preprint arXiv:1711.11585},
  year={2017}
}
@article{chen2018high,
  title={High Resolution Face Completion with Multiple Controllable Attributes via Fully End-to-End Progressive Generative Adversarial Networks},
  author={Chen, Zeyuan and Nie, Shaoliang and Wu, Tianfu and Healey, Christopher G},
  journal={arXiv preprint arXiv:1801.07632},
  year={2018}
}
@article{odena2016conditional,
  title={Conditional image synthesis with auxiliary classifier gans},
  author={Odena, Augustus and Olah, Christopher and Shlens, Jonathon},
  journal={arXiv preprint arXiv:1610.09585},
  year={2016}
}
@inproceedings{chongxuan2017triple,
  title={Triple generative adversarial nets},
  author={Chongxuan, LI and Xu, Taufik and Zhu, Jun and Zhang, Bo},
  booktitle={Advances in Neural Information Processing Systems},
  pages={4091--4101},
  year={2017}
}
@inproceedings{tran2017bayesian,
  title={A Bayesian Data Augmentation Approach for Learning Deep Models},
  author={Tran, Toan and Pham, Trung and Carneiro, Gustavo and Palmer, Lyle and Reid, Ian},
  booktitle={NIPS},
  pages={2794--2803},
  year={2017}
}
@article{salehinejad2017generalization,
  title={Generalization of Deep Neural Networks for Chest Pathology Classification in X-Rays Using Generative Adversarial Networks},
  author={Salehinejad, Hojjat and Valaee, Shahrokh and Dowdell, Tim and Colak, Errol and Barfett, Joseph},
  journal={arXiv preprint arXiv:1712.01636},
  year={2017}
}
@article{guibas2017synthetic,
  title={Synthetic Medical Images from Dual Generative Adversarial Networks},
  author={Guibas, John T and Virdi, Tejpal S and Li, Peter S},
  journal={arXiv preprint arXiv:1709.01872},
  year={2017}
}
@inproceedings{fawzi2016adaptive,
  title={Adaptive data augmentation for image classification},
  author={Fawzi, Alhussein and Samulowitz, Horst and Turaga, Deepak and Frossard, Pascal},
  booktitle={Image Processing (ICIP), 2016 IEEE International Conference on},
  pages={3688--3692},
  year={2016},
  organization={Ieee}
}

@article{shen2017end,
  title={End-to-end Training for Whole Image Breast Cancer Diagnosis using An All Convolutional Design},
  author={Shen, Li},
  journal={arXiv preprint arXiv:1708.09427},
  year={2017}
}
@misc{hinton2012neural,
  title={Neural networks for machine learning lecture 6a overview of mini-batch gradient descent},
  author={Hinton, Geoffrey and Srivastava, Nitish and Swersky, Kevin}
}
@inproceedings{dhungel2017fully,
  title={Fully automated classification of mammograms using deep residual neural networks},
  author={Dhungel, Neeraj and Carneiro, Gustavo and Bradley, Andrew P},
  booktitle={Biomedical Imaging (ISBI 2017), 2017 IEEE 14th International Symposium on},
  pages={310--314},
  year={2017},
  organization={IEEE}
}
@article{donahue2016adversarial,
  title={Adversarial feature learning},
  author={Donahue, Jeff and Kr{\"a}henb{\"u}hl, Philipp and Darrell, Trevor},
  journal={arXiv preprint arXiv:1605.09782},
  year={2016}
}
@inproceedings{liu2007generative,
  title={Generative Oversampling for Mining Imbalanced Datasets.},
  author={Liu, Alexander and Ghosh, Joydeep and Martin, Cheryl E}
}
@inproceedings{forestier2017generating,
  title={Generating synthetic time series to augment sparse datasets},
  author={Forestier, Germain and Petitjean, Fran{\c{c}}ois and Dau, Hoang Anh and Webb, Geoffrey I and Keogh, Eamonn},
  booktitle={Data Mining (ICDM), 2017 IEEE International Conference on},
  pages={865--870},
  year={2017},
  organization={IEEE}
}
@article{douzas2018effective,
  title={Effective data generation for imbalanced learning using conditional generative adversarial networks},
  author={Douzas, Georgios and Bacao, Fernando},
  journal={Expert Systems with Applications},
  volume={91},
  pages={464--471},
  year={2018},
  publisher={Elsevier}
}
@article{hwang2017disease,
  title={Disease Prediction from Electronic Health Records Using Generative Adversarial Networks},
  author={Hwang, Uiwon and Choi, Sungwoon and Yoon, Sungroh},
  journal={arXiv preprint arXiv:1711.04126},
  year={2017}
}
@article{chawla2002smote,
  title={SMOTE: synthetic minority over-sampling technique},
  author={Chawla, Nitesh V and Bowyer, Kevin W and Hall, Lawrence O and Kegelmeyer, W Philip},
  journal={Journal of artificial intelligence research},
  volume={16},
  pages={321--357},
  year={2002}
}
@misc{zhu2017data,
Author = {Xinyue Zhu and Yifan Liu and Zengchang Qin and Jiahong Li},
Title = {Data Augmentation in Emotion Classification Using Generative Adversarial Networks},
Year = {2017},
Eprint = {arXiv:1711.00648},
}
@article{wilson1972asymptotic,
  title={Asymptotic properties of nearest neighbor rules using edited data},
  author={Wilson, Dennis L},
  journal={IEEE Transactions on Systems, Man, and Cybernetics},
  number={3},
  pages={408--421},
  year={1972},
  publisher={IEEE}
}
@article{douzas2017geometric,
  title={Geometric SMOTE: Effective oversampling for imbalanced learning through a geometric extension of SMOTE},
  author={Douzas, Georgios and Bacao, Fernando},
  journal={arXiv preprint arXiv:1709.07377},
  year={2017}
}
@article{kingma2014adam,
  title={Adam: A method for stochastic optimization},
  author={Kingma, Diederik P and Ba, Jimmy},
  journal={arXiv preprint arXiv:1412.6980},
  year={2014}
}
@inproceedings{han2005borderline,
  title={Borderline-SMOTE: a new over-sampling method in imbalanced data sets learning},
  author={Han, Hui and Wang, Wen-Yuan and Mao, Bing-Huan},
  booktitle={International Conference on Intelligent Computing},
  pages={878--887},
  year={2005},
  organization={Springer}
}
@inproceedings{brain1999effect,
  title={On the effect of data set size on bias and variance in classification learning},
  author={Brain, Damien and Webb, G}
}
@inproceedings{wong2016understanding,
  title={Understanding data augmentation for classification: when to warp?},
  author={Wong, Sebastien C and Gatt, Adam and Stamatescu, Victor and McDonnell, Mark D},
  booktitle={Digital Image Computing: Techniques and Applications (DICTA), 2016 International Conference on},
  pages={1--6},
  year={2016},
  organization={IEEE}
}
@article{fallahi2011expert,
  title={An expert system for detection of breast cancer using data preprocessing and bayesian network},
  author={Fallahi, Amir and Jafari, Shahram},
  journal={International Journal of Advanced Science and Technology},
  volume={34},
  pages={65--70},
  year={2011}
}
@article{lecun1998gradient,
  title={Gradient-based learning applied to document recognition},
  author={LeCun, Yann and Bottou, L{\'e}on and Bengio, Yoshua and Haffner, Patrick},
  journal={Proceedings of the IEEE},
  volume={86},
  number={11},
  pages={2278--2324},
  year={1998},
  publisher={IEEE}
}
@article{krizhevsky2009learning,
  title={Learning multiple layers of features from tiny images},
  author={Krizhevsky, Alex},
  year={2009},
  publisher={Citeseer}
}
@article{simonyan2014very,
  title={Very deep convolutional networks for large-scale image recognition},
  author={Simonyan, Karen and Zisserman, Andrew},
  journal={arXiv preprint arXiv:1409.1556},
  year={2014}
}
@article{arjovsky2017wasserstein,
  title={Wasserstein gan},
  author={Arjovsky, Martin and Chintala, Soumith and Bottou, L{\'e}on},
  journal={arXiv preprint arXiv:1701.07875},
  year={2017}
}
@misc{openai_2018, title={Requests for Research 2.0}, url={https://blog.openai.com/requests-for-research-2/}, journal={OpenAI Blog}, publisher={OpenAI Blog}, author={OpenAI}, year={2018}, month={Feb}}
@article{sixt2016rendergan,
  title={Rendergan: Generating realistic labeled data},
  author={Sixt, Leon and Wild, Benjamin and Landgraf, Tim},
  journal={arXiv preprint arXiv:1611.01331},
  year={2016}
}
@article{devries2017dataset,
  title={Dataset augmentation in feature space},
  author={DeVries, Terrance and Taylor, Graham W},
  journal={arXiv preprint arXiv:1702.05538},
  year={2017}
}

@article{curto2017high,
  title={High-Resolution Deep Convolutional Generative Adversarial Networks},
  author={Curto, Joachim D and Zarza, Irene C and De La Torre, Fernando and King, Irwin and Lyu, Michael R},
  journal={arXiv preprint arXiv:1711.06491},
  year={2017}
}

@article{lotter2017multi,
  title={A Multi-Scale CNN and Curriculum Learning Strategy for Mammogram Classification},
  author={Lotter, William and Sorensen, Greg and Cox, David},
  journal={arXiv preprint arXiv:1707.06978},
  year={2017}
}

@inproceedings{salimans2016improved,
  title={Improved techniques for training gans},
  author={Salimans, Tim and Goodfellow, Ian and Zaremba, Wojciech and Cheung, Vicki and Radford, Alec and Chen, Xi},
  booktitle={NIPS},
  pages={2234--2242},
  year={2016}
}

@inproceedings{gulrajani2017improved,
  title={Improved training of wasserstein gans},
  author={Gulrajani, Ishaan and Ahmed, Faruk and Arjovsky, Martin and Dumoulin, Vincent and Courville, Aaron C},
  booktitle={NIPS},
  pages={5769--5779},
  year={2017}
}

@article{berthelot2017began,
  title={Began: Boundary equilibrium generative adversarial networks},
  author={Berthelot, David and Schumm, Tom and Metz, Luke},
  journal={arXiv preprint arXiv:1703.10717},
  year={2017}
}

@article{kodali2017train,
  title={How to train your DRAGAN},
  author={Kodali, Naveen and Abernethy, Jacob and Hays, James and Kira, Zsolt},
  journal={arXiv preprint arXiv:1705.07215},
  year={2017}
}
@article{antoniou2017data,
  title={Data Augmentation Generative Adversarial Networks},
  author={Antoniou, Antreas and Storkey, Amos and Edwards, Harrison},
  journal={arXiv preprint arXiv:1711.04340},
  year={2017}
}
@inproceedings{gulrajani2017improved,
  title={Improved training of wasserstein gans},
  author={Gulrajani, Ishaan and Ahmed, Faruk and Arjovsky, Martin and Dumoulin, Vincent and Courville, Aaron C},
  booktitle={NIPS},
  pages={5769--5779},
  year={2017}
}
  
\end{document}